\documentclass{article}

\usepackage{microtype}
\usepackage{graphicx}
\usepackage{subfigure}
\usepackage{booktabs} % for professional tables

\usepackage{amsmath}
\usepackage{amssymb}
\usepackage{amsthm}
\usepackage[inline]{enumitem}
\usepackage[utf8]{inputenc}
\DeclareUnicodeCharacter{00B1}{\pm}

\newtheorem{theorem}{Theorem}[section]

\newtheorem{lemma}[theorem]{Lemma}

\newcommand{\beginsupplement}{%
        \setcounter{table}{0}
        \renewcommand{\thetable}{S\arabic{table}}%
        \setcounter{figure}{0}
        \renewcommand{\thefigure}{S\arabic{figure}}%
}

\usepackage{hyperref}

\usepackage[accepted]{icml2020}

\icmltitlerunning{TrajectoryNet: A Dynamic Optimal Transport Network for Modeling Cellular Dynamics}

\begin{document}

\twocolumn[
\icmltitle{TrajectoryNet: A Dynamic Optimal Transport Network for Modeling Cellular Dynamics }



\icmlsetsymbol{equal}{*}

%\begin{icmlauthorlist}
%\icmlauthor{Alexander Tong}{cs}
%\icmlauthor{Jessie Huang}{cs}
%\icmlauthor{Guy Wolf}{equal,dms,mila}
%\icmlauthor{David van Dijk}{equal,cardio,cs}
%\icmlauthor{Smita Krishnaswamy}{equal,gene,cs}
%\end{icmlauthorlist}

%\icmlaffiliation{cs}{Department of Computer Science, Yale University, New Haven, CT, USA}
%\icmlaffiliation{dms}{Department of Mathematics \& Statistics, Universit\'{e} de Montr\'{e}al, Montr\'{e}al, QC, Canada}
%\icmlaffiliation{mila}{Mila -- Quebec AI Institute, Montr\'{e}al, QC, Canada}
%\icmlaffiliation{cardio}{Department of Cardiology, Yale University, New Haven, CT, USA}
%\icmlaffiliation{gene}{Department of Genetics, Yale University, New Haven, CT, USA}

%\icmlcorrespondingauthor{Guy Wolf}{guy.wolf@umontreal.ca}
%\icmlcorrespondingauthor{David van Dijk}{david.vandijk@yale.edu}
%\icmlcorrespondingauthor{Smita Krishnaswamy}{smita.krishnaswamy@yale.edu}

\begin{icmlauthorlist}
\icmlauthor{Alexander Tong}{cs}
\icmlauthor{Jessie Huang}{cs}
\icmlauthor{Guy Wolf}{equal,dms,mila}
\icmlauthor{David van Dijk}{equal,cs,cardio}
\icmlauthor{Smita Krishnaswamy}{equal,cs,gene}
\end{icmlauthorlist}

\icmlaffiliation{cs}{Department of Computer Science, Yale University, New Haven, CT, USA}
\icmlaffiliation{dms}{Department of Mathematics \& Statistics, Universit\'{e} de Montr\'{e}al, Montr\'{e}al, QC, Canada}
\icmlaffiliation{mila}{Mila -- Quebec AI Institute, Montr\'{e}al, QC, Canada}
\icmlaffiliation{cardio}{Internal Medicine, Cardiology, Yale University, New Haven, CT, USA}
\icmlaffiliation{gene}{Department of Genetics, Yale University, New Haven, CT, USA}

%\icmlcorrespondingauthor{Guy Wolf}{guy.wolf@umontreal.ca}
%\icmlcorrespondingauthor{David van Dijk}{david.vandijk@yale.edu}
\icmlcorrespondingauthor{Smita Krishnaswamy}{smita.krishnaswamy@yale.edu}

\icmlkeywords{Dynamical systems, differential equations, neural network}

\vskip 0.3in
]



\printAffiliationsAndNotice{\icmlEqualContribution} % otherwise use the standard text.

\begin{abstract}
%It is increasingly common to encounter data in the form of cross-sectional population measurements over time, particularly in biomedical settings. Recent attempts to model individual trajectories from this data use optimal transport to create pairwise matchings between time points. However, these methods cannot model non-linear paths common in many underlying dynamic systems. We establish a link between continuous normalizing flows and dynamic optimal transport to model the expected paths of points over time. Continuous normalizing flows are generally under constrained, as they are allowed to take an arbitrary path from the source to the target distribution. We present {\em TrajectoryNet}, which controls the continuous paths taken between distributions. We show how this is particularly applicable for studying cellular dynamics in data from single-cell RNA sequencing (scRNA-seq) technologies, and that TrajectoryNet improves upon recently proposed static optimal transport-based models that can be used for interpolating cellular distributions.

%It is increasingly common to encounter data from dynamic processes in the form of static cross-sectional population measurements over time, particularly in biomedical settings.
It is increasingly common to encounter data from dynamic processes captured by static cross-sectional measurements over time, particularly in biomedical settings. Recent attempts to model individual trajectories from this data use optimal transport to create pairwise matchings between time points. However, these methods cannot model continuous dynamics and non-linear paths that entities can take in these systems. To address this issue, we establish a link between continuous normalizing flows and dynamic optimal transport, that allows us to model the expected paths of points over time. Continuous normalizing flows are generally under constrained, as they are allowed to take an arbitrary path from the source to the target distribution. We present {\em TrajectoryNet}, which controls the continuous paths taken between distributions to produce dynamic optimal transport. We show how this is particularly applicable for studying cellular dynamics in data from single-cell RNA sequencing (scRNA-seq) technologies, and that TrajectoryNet improves upon recently proposed static optimal transport-based models that can be used for interpolating cellular distributions.
\end{abstract}

\section{Introduction}
    In data science we are often confronted with cross-sectional samples of time-varying phenomena, especially in biomedical data. Examples include health measurements of different age cohorts~\cite{oeppen_broken_2002}, or disease measurements at different stages of disease progression~\cite{waddington_epigenotype_1942}. In these measurements we consider data that is sampled at multiple timepoints, but at each timepoint we have access only to a distribution (cross-section) of the population at that time. Extracting the longitudinal dynamics of development or disease from static snapshot measurements can be challenging as there are few methods of interpolation between distributions. Further exacerbating this problem is the fact that the same entities are often not measured at each time, resulting in a lack of point-to-point correspondences. Here, we propose to formulate this problem as one of unbalanced dynamic transport, where the goal is to transport entities from one cross sectional measurement to the next using efficient and smooth paths.  Our main contribution is to establish a link between {\em continuous normalizing flows} (CNF)~\cite{grathwohl_ffjord:_2019} and dynamic optimal transport~\cite{benamou_computational_2000}, allowing us to efficiently solve the transport problem using a Neural ODE framework~\cite{chen_neural_2018}. To our knowledge, TrajectoryNet\footnote{Code is available here: \url{https://github.com/KrishnaswamyLab/TrajectoryNet}} is the first method to consider the specific paths taken by a CNF between distributions.

% New paragraph
The continuous normalizing flow formulation allows us to generalize optimal transport to a series of distributions as in recent work~\cite{chen_measure-valued_2018, benamou_second-order_2019}. These works focus on the theoretical aspects of the problem, here focus on the computational aspects. This link allows us to smooth flows over multiple and possibly unevenly spaced distributions in high dimensions. This matches the setting of time series data from single-cell RNA sequencing.

Single-cell RNA sequencing~\cite{macosko_highly_2015} is a relatively new technology that has made it possible for scientists to randomly sample the entire transcriptome, i.e., 20-30 thousand species of mRNA molecules representing transcribed genes of the cell. This technology can reveal detailed information about the identity of individual cells based on transcription factors, surface marker expression, cell cycle and many other facets of cellular behavior. In particular, this technology can be used to learn how cells differentiate from one state to another: for example, from embryonic stem cells to specified lineages such as neuronal or cardiac. However, hampering this understanding is the fact that scRNA-seq only offers static snapshots of data, since all cells are destroyed upon measurement. Thus it is impossible to monitor how an individual cell changes over time. Moreover, due to the expensive nature of this technology, generally only a handful of discrete timepoints are collected in measuring any transition process. TrajectoryNet is especially well suited to this data modality. Existing methods attempt to infer a trajectory within one timepoint~\cite{haghverdi_diffusion_2016, saelens_comparison_2019, la_manno_rna_2018}, or interpolate linearly between two timepoints~\cite{yang_scalable_2019, schiebinger_optimal-transport_2019}, but TrajectoryNet can interpolate non-linearly using information from more than two timepoints. TrajectoryNet has advantages over existing methods in that it: 

\begin{enumerate}
\item can interpolate by following the manifold of observed entities between measured timepoints, thereby solving the static-snapshot problem, 
\item can create continuous-time trajectories of individual entities, giving researchers the ability to follow an entity in time,  
\item forms a deep representational model of system dynamics, which can then be used to understand drivers of dynamics (gene logic in the cellular context), via perturbation of this deep model.
\end{enumerate}

While our experiments apply this work specifically to cellular dynamics, these penalties can be used in many other situations where we would like to model dynamics based on cross-sectional population level data.

%Finally, the most interesting insights into differentiation are gene regulatory logic that creates state change, and this has proven very difficult to learn using association-based methods (such as mutual information), due noisy measurements and lack of ability to infer causality. 

%We note that this does not get around the sample complexity for general empirical distributions \todo{cite} but is not 

%When modeling a continuous time dynamical system using distributional measurements as in scRNA-seq our prior knowledge about the dynamics can most easily be encoded in terms of the paths taken by individual cells. 

    \begin{figure*}[th]
    \begin{center}
    \includegraphics[width=1 \linewidth]{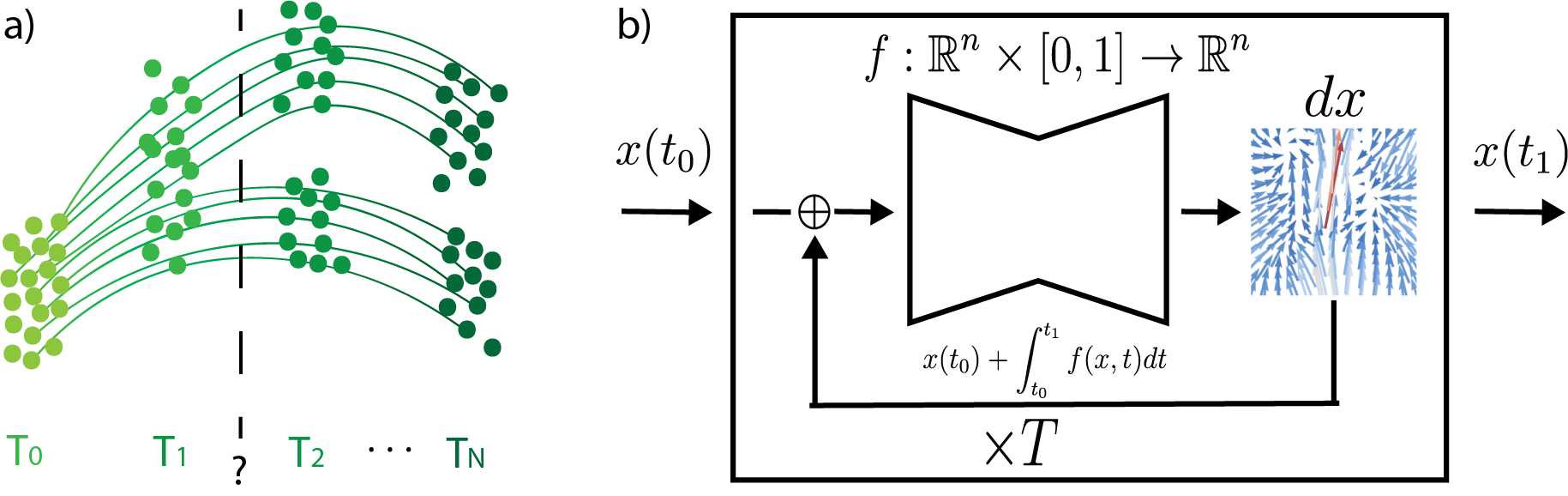}
    \end{center}
    %\vspace{-5mm}
    \caption{TrajectoryNet learns trajectories of particles from distributions sampled over time. We use a Neural ODE to learn the derivative of the dynamics function. To find the output at time $t_1$ for a given input at time $t_0$ we integrate $T$ times letting the ODE solver choose the integration timepoints.}
    \label{fig:Tnet_Arch}
\end{figure*}
\section{Background and Related Work}
    \paragraph{Optimal Transport.} Introduced originally by \cite{monge_memoire_1781} and in modern form by \cite{kantorovich_translocation_1942}, the linear program formulation of static optimal transport (OT) has the relatively high cost of $O(n^3)$ for discrete measures. Recently, there have been a number of fast approximations using entropic regularization. \citet{cuturi_sinkhorn_2013} presented a parallel algorithm for the discrete case as an application of Sinkhorn's algorithm \cite{sinkhorn_relationship_1964}. Recent effort approximates OT on subspaces \cite{muzellec_subspace_2019} or even a single dimension \cite{kolouri_generalized_2019}. These efforts emphasis the importance to the field of obtaining fast OT algorithms. Another direction that has recently received increased attention is in unbalanced optimal transport where the goal is to relax the problem to add and remove mass~\cite{benamou_numerical_2003, chizat_unbalanced_2018, liero_optimal_2018, schiebinger_optimal-transport_2019}. While many efficient static optimal transport algorithms exist, and recently for the unbalanced case~\cite{yang_scalable_2019}, much less attention has focused on dynamic optimal transport, the focus of this work. 

%as OT has found uses in diverse fields such as computer graphics~\cite{bonneel_displacement_2011}, medical imaging~\cite{basu_detecting_2014},  and \todo{MORE}. 

\paragraph{Dynamic Optimal Transport.} Another formulation of optimal transport is known as {\em dynamic} optimal transport. \citet{benamou_computational_2000} showed how the addition of a natural time interpolation variable gives an alternative interpretation with links to fluid dynamics that surprisingly leads to a convex optimization problem. However, while solvers for the discretized dynamic OT problem are effective in low dimensions and for small problems they require a discretization of space into grids giving cost exponential in the dimension (See \citet{peyre_computational_2019} Chap. 7 for a good overview of this problem). One of our main contributions is to provide an approximate solver for high dimensional smooth problems using a neural network.

%Both static and dynamic forms of OT can be interpreted as optimizing over a paths space. In both cases the distribution of endpoints of paths are constrained to match the source and target distributions, however the static formulation ca be thought of as optimizing over all possible sets of linear paths, where the dynamic formulation optimizes over curves. A relaxation to optimization over curves and an approximate match between the distribution of curve endpoints and the target distribution allows a natural neural network solution. 

\paragraph{Single-cell Trajectories from a Static Snapshot.} Temporal interpolation in single-cell data started with solutions that attempt to infer an axis within one single time point of data cell ``pseudotime" -- used as a proxy for developmental progression -- using known markers of development and the asynchronous nature of cell development~\cite{trapnell_dynamics_2014, bendall_single-cell_2014}. An extensive comparison of 45 methods for this type of analysis gives method recommendations based on prior assumptions on the general structure of the data~\cite{saelens_comparison_2019}. However, these methods can be biased and fail in a number of circumstances~\cite{weinreb_fundamental_2018, lederer_emergence_2020} and do not take into account experimental time. 

\paragraph{Matching Populations from Multiple Time Points.}  Recent methods get around some of these challenges using multiple timepoints~\cite{hashimoto_learning_2016, schiebinger_optimal-transport_2019, yang_scalable_2019}. However, these methods generally resort to matching populations between coarse-grained timepoints, but do not give much insight into how they move between measured timepoints. Often paths are assumed to minimize total Euclidean cost, which is not realistic in this setting. In contrast, the methods that estimate dynamics from single timepoints ~\cite{la_manno_rna_2018, bergen_generalizing_2019, erhard_scslam-seq_2019, hendriks_nasc-seq_2019} have the potential to give relatively accurate estimation of local direction, but cannot give accurate global estimation of distributional shift. A recent line of work on generalizing splines to distributions \cite{chen_measure-valued_2018, benamou_second-order_2019} investigates this problem from a theoretical perspective, but provides no efficient implementation.

With TrajectoryNet, we aim to unite these approaches into a single model combining in inferring continuous time trajectories from multiple timepoints, globally, while respecting local dynamics within a single timepoint. 

\section{Preliminaries}
    We provide an overview of static optimal transport, dynamic optimal transport~\cite{benamou_computational_2000}, and continuous normalizing flows.

\subsection{The Monge-Kantorovich Problem}
We 
adopt notation from the standard text~\cite{villani_optimal_2008}. For two probability measures $\mu, \nu$ defined on $\mathcal{X} \subset \mathbb{R}^n$, let $\Pi(\mu, \nu)$ denote the set of all joint probability measures on $\mathcal{X} \times \mathcal{X}$ whose marginals are $\mu$ and $\nu$. Then the p-Kantorovich  distance (or Wasserstein distance of order $p$)between $\mu$ and $\nu$ is
\begin{equation}
    W(\mu, \nu)_p := \biggl (\inf_{\pi \in \Pi(\mu, \nu)} \int_{\mathcal{X} \times \mathcal{Y}} d(x, y) d\pi(x,y) \biggr ) ^ {1/p},
\end{equation}

where $p\in [1,\infty)$. This formulation has led to many useful interpretations both in GANs and biological networks. For the entropy regularized problem, the Sinkhorn algorithm~\cite{sinkhorn_relationship_1964} provides a fast and parallelizable numerical solution in the discrete case. Recent work tackles computationally efficient solutions to the exact problem~\cite{jambulapati_2019} for the discrete case. However, for the continuous case solutions to the discrete problem in high dimensional spaces do not scale well. As the rate of convergence of the empirical Wasserstein metric between empirical measures $\hat{\mu}$ and $\hat{\nu}$ with bounded support is shown in \cite{dudley_speed_1969} to be
\begin{equation}
    \mathbb{E}[|W_p(\hat{\mu}_n, \hat{\nu}_n) - W_p(\mu, \nu)|] = O(n^{-\frac{1}{d}})
\end{equation}
where $d$ is the ambient dimension. However, recent work shows that in high dimensions a more careful treatment that the rate depends on the intrinsic dimension not the ambient dimension~\cite{weed_sharp_2019}. As long as data lies in a low dimensional manifold in ambient space, then we can reasonably approximate the Wasserstein distance. In this work we approximate the support of this manifold using a neural network.

\subsection{Dynamic Optimal Transport}
\citet{benamou_computational_2000} defined and explored a dynamic version of Kantorovich distance. Their work linked optimal transport distances with dynamics and partial differential equations (PDEs). For a fixed time interval $[t_0, t_1]$ with smooth enough, time dependent density and velocity fields, $P(x,t) \ge 0, f(x,t) \in \mathbb{R}^d$, subject to the continuity equation
\begin{equation}\label{eq:benamou:c1}
    \partial_t P + \nabla \cdot (P f) = 0
\end{equation}
for $t_0 < t < t_1$ and $x \in \mathbb{R}^d$, and the conditions
\begin{equation}\label{eq:benamou:c2}
    P(\cdot, t_0) = \mu, \quad P(\cdot, t_1) = \nu
\end{equation}
we can relate the squared $L^2$ Wasserstein distance to $(P, f)$ in the following way
\begin{equation}\label{eq:benamou}
    W(\mu, \nu)^2_2 = \inf_{(P,f)} (t_1 - t_0) \int_{\mathbb{R}^d} \int_{t_0}^{t_1} P(x,t) |f(x,t)|^2 dt dx
\end{equation}
In other words, a velocity field $f(x,t)$ with minimum $L^2$ norm that transports mass at $\mu$ to mass at $\nu$ when integrated over the time interval is the optimal plan for an $L^2$ Wasserstein distance. The continuity equation is applied over all points of the field and asserts that no point is a source or sink for mass. The solution to this flow can be shown to follow a pressureless flow on a time-dependent potential function. Mass moves with constant velocity that linearly interpolates between the initial and final measures. For problems where interpolation of the density between two known densities is of interest, this formulation is very attractive. Existing computational methods for solving the dynamic formulation for continuous measures approximate the flow using a discretization of space-time~\cite{papadakis_optimal_2014}. This works well in low dimensions, but scales poorly to high dimensions as the complexity is exponential in the input dimension $d$. We next give background on continuous normalizing flows, which we show can provide a solution with computational complexity polynomial in $d$.

\subsection{Continuous Normalizing Flows}
A normalizing flow~\cite{rezende_variational_2015} transforms a parametric (usually simple) distribution to a more complicated one. Using an invertible transformation $f$ applied to an initial latent random variable $y$ with density $P_y$, We define $x=f(y)$ as the output of the flow. Then by the change of variables formula, we can compute the density of the output $x$:

\begin{equation}
    \log P_x(\cdot) = \log P_y(\cdot) - \log \left | \det \frac{\partial f}{\partial y} \right |
\end{equation}

A large effort has gone into creating architectures where the log determinant of the Jacobian is efficient to compute~\cite{rezende_variational_2015,kingma_improving_2016,papamakarios_masked_2017}.

Now consider a continuous-time transformation, where the derivative of the transformation is parameterized by $\theta$, thus at any timepoint $t$, $\frac{\partial x(t)}{\partial t} = f_\theta(x(t),t)$. At the initial time $t_0$, $x(t_0)$ is drawn from a distribution $P(x, t_0)$ which we also denote $P_{t_0}(x)$ for clarity, and it's continuously transformed to $x(t_1)$ by following the differential equation $f_\theta(x(t),t)$: 
\begin{align}\label{eq:cnf_integration}
    &x(t_1) = x(t_0) + \int_{t_0}^{t_1} f_\theta(x(t), t) dt, \quad
    x(t_0) \sim P_{t_0}(x), \nonumber \\
    &\log P_{t_1}(x(t_1)) = \nonumber \\
    &\quad\quad \log P_{t_0}(x(t_0)) - \int_{t_0}^{t_1} Tr \left ( \frac{\partial f_\theta(x(t),t)}{\partial x(t)} \right ) dt,
\end{align}
where at any time $t$ associated with every $x$ through the flow can be found by following the inverse flow. This model is referred as continuous normalizing flows (CNFs)~\cite{chen_neural_2018}. It can be likened to the dynamic version of optimal transport, where we model the measure over time rather than the mapping from $P_{t_0}$ to $P_{t_1}$

Unsurprisingly, there is a deep connection between CNFs and dynamic optimal transport. In the next section we exploit this connection and show how CNFs can be used to provide a high dimensional solution to the dynamic optimal transport problem with TrajectoryNet. 

    \begin{figure}[ht]
    \begin{center}
    \includegraphics[width=1 \linewidth]{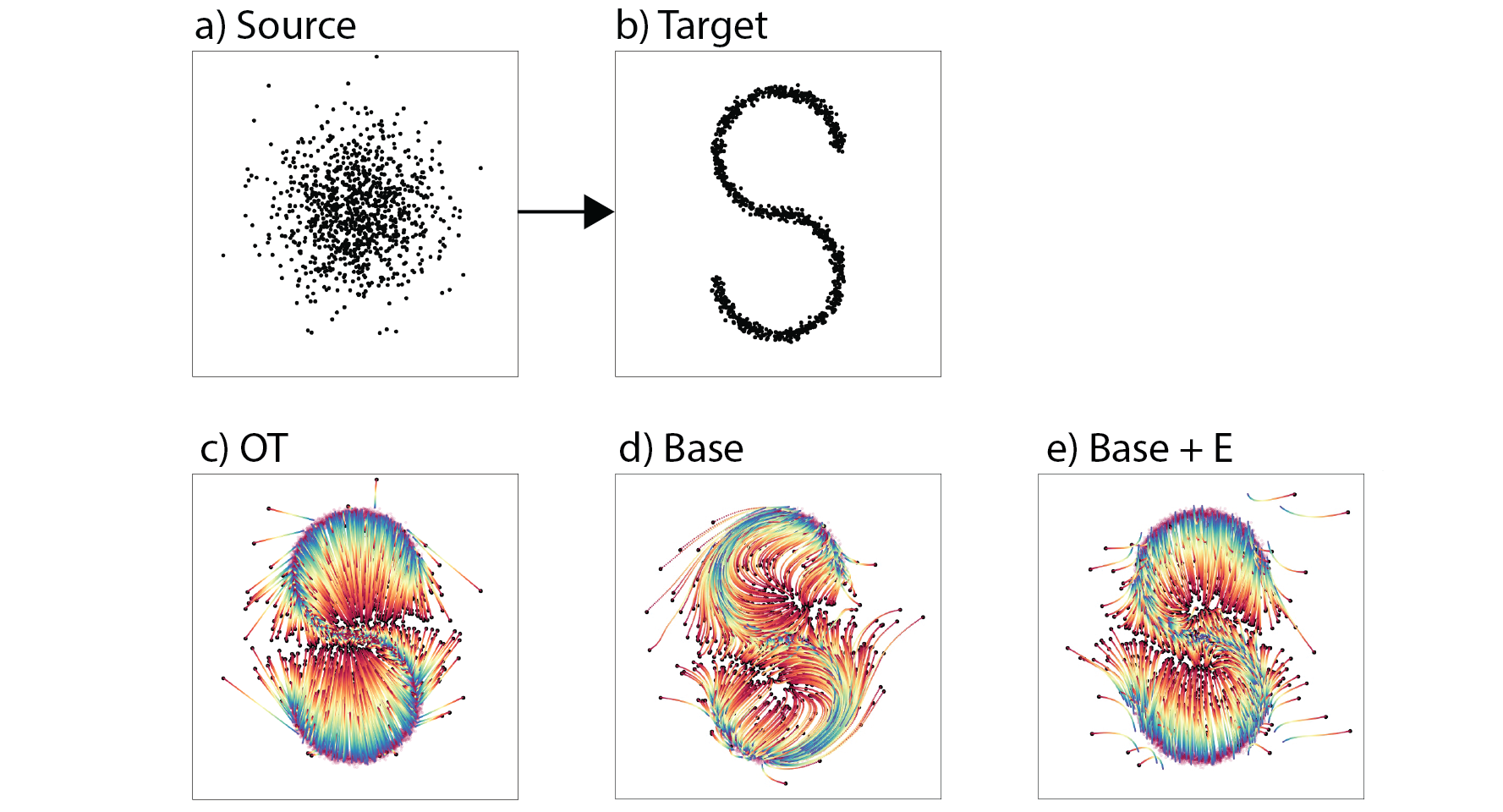}
    \end{center}
    %\vspace{-5mm}
    \caption{Transporting a Gaussian (a) to an S-curve (b) via (c) static optimal transport, (d) Base TrajectoryNet without regularization follows density (e) TrajectoryNet with energy regularization demonstrates more straight paths similar to OT. }
    %\caption{Our model approximates dynamic optimal transport by following low kinetic energy paths.}
    \label{fig:scurve}
\end{figure}
\section{TrajectoryNet: Efficient Dynamic Optimal Transport}
    In this section, we first describe how to adapt continuous normalizing flows to approximate dynamic optimal transport in (Section \ref{sec:energy}). We then describe further adaptations for analysis of single-cell data in (Section \ref{sec:single-cell-data}) and finally provide training details in (Section \ref{sec:training}).

%For a dataset with measured samples at timepoints $\{t_1, t_2, \ldots, t_k\}$ we denote the measured observations from each timepoint as $\{\mathcal{X}_{t_1}, \mathcal{X}_{t_2}, \ldots, \mathcal{X}_{t_k}\}$. The goal of TrajectoryNet is to create a continuous mapping between these timepoints. To this end, we train a neural network with parameters $\theta$ that models the instantaneous change of a cell state over time $\frac{\partial x(t)}{\partial t} = f_\theta(x,t)$. TrajectoryNet predicts the instantaneous next state of an individual cell, but is trained on a series of distributions.

\subsection{Dynamic OT Approximation via Regularized CNF}\label{sec:energy} Continuous normalizing flows use a maximum likelihood objective which can be equivalently expressed as a KL divergence. In TrajectoryNet we add an energy regularization to approximate dynamic OT. Dynamic OT is expressed with an optimization over flows with constraints at $t_0$ and $t_1$ (see eq.~\eqref{eq:benamou:c2}). By relaxing this constraint to minimizing a divergence at $t_1$ CNFs can approximate dynamic OT.

For sufficiently large $\lambda$ under constraint \eqref{eq:benamou:c1} this converges to the optimal solution in \eqref{eq:benamou}. This is encapsulated in the following theorem. See Appendix~\ref{app:proof} for proof.
\begin{theorem}\label{thm:approach}
With time varying field $f(x,t): \mathbb{R}^d \times \mathbb{R} \rightarrow \mathbb{R}^d$ and density $P(x,t): \mathbb{R}^d \times \mathbb{R} \rightarrow \mathbb{R^+}$ such that $\int P(x,t) dx = 1$ for all $t_0 \le t \le t_1$ and subject to the continuity \eqref{eq:benamou:c1}. There exists a sufficiently large $\lambda$ such that
\begin{align}\label{eq:wasserstein}
    W(\mu, \nu)^2_2 &=(t_1 - t_0) \inf_{(P,f)} \mathop{\mathbb{E}}_{x_0 \sim \mu} \left [ \int_{t_0}^{t_1} \left \| f(x(t),t) \right \|^2 dt \right ] \nonumber \\
    &+ \lambda \text{KL}(P(\cdot, t_1) \parallel \nu); \nonumber
     \quad \text{s.t. } P(\cdot, t_0) = \mu
\end{align}
\end{theorem}

Intuitively, a continuous normalizing flow with a correctly scaled penalty on the squared norm of $f$ approximates the $W_2$ transport between $\mu$ and $\nu$. Dynamic optimal transport can be thought of as finding a distribution over paths such that the beginnings of the paths match the source distribution, end of the path matches the target distribution, and the cost of the transport is measured by expected path length. Continuous normalizing flows relax the target distribution match with a KL-divergence penalty. When the KL-divergence is small then the constraint is satisfied. However, CNFs usually do not enforce the path length constraint, which we add using a penalty on the norm of $f$ as defined by the Neural ODE. When we impose this penalty over uniformly sampled data, this is equivalent to penalizing the expected path length.

We can approximate the first part of this continuous time equation using a Riemann sum as $\mathop{\mathbb{E}}_{x_0 \sim \mu} \left [ \int_{t_0}^{t_1}  \left \| f_\theta (x,t)\right \|^2 dt \right ] = \sum_{x \sim P_{t_0}} \sum_{i = t_0}^{t_1} \Delta t_i \left \|f_\theta(x,t) \right \|^2$. This requires a forward integration using a standard ODE solver to compute as shown in \citet{chen_neural_2018}. If we consider the case where the divergence is small, then this can be combined with the standard backwards pass for even less added computation. Instead of penalizing $\left \| f_\theta (x,t)\right \|^2$ on a forward pass, we penalize the same quantity on a backwards pass. Using the maximum likelihood and KL divergence equivalence, we obtain a loss 
\begin{equation}
    L(x) = - \log p(x) + \lambda_e \int_t \left \| f(x(t),t) \right \|^2
\end{equation}
Where the integral above is computed using an ODE solver. In practice, both a penalty on the Jacobian or additional training noise helped to get straight paths with a lower energy regularization $\lambda_e$. We found that a value of $\lambda_e$ large enough to encourage straight paths, unsurprisingly also shortens the paths undershooting the target distribution. To counteract this, we add a penalty on the norm of the Jacobian of $f$ as used in~\citet{vincent_stacked_2010, rifai_contractive_2011-1}. Since $f$ represents the derivative of the path, this discourages paths with high local curvature, and can be thought of as penalizing the second derivative (acceleration) of the flow. Our energy loss is then
\begin{equation}
    L_{energy}(x) = \lambda_e \int_t \left \| f(\tilde{x},t) \right \|^2 + \lambda_j \int_t \|J_f(\tilde{x})\|^2_F ,
\end{equation}
where $\|J_f(x)\|^2_F$ is the Frobenius norm of the Jacobian of $f$. Comparing Figure~\ref{fig:scurve}(e) to (d) we demonstrate the effect of this regularization. Without energy regularization TrajectoryNet paths follow the data. However, with energy regularization we approach the paths of the optimal map. TrajectoryNet solution biases towards undershooting the target distribution. Our energy loss gives control over how much to penalize indirect, high energy paths.

Optimal transport is traditionally performed between a source and target distribution. Extensions to a series of distributions is normally done by performing optimal transport between successive pairs of distributions as in~\citet{schiebinger_optimal-transport_2019}. This creates flows that have discontinuities at the sampled times, which may be undesirable when the underlying system is smooth in time as in biological systems. The dynamic model approximates dynamic OT for two timepoints, but by using a single smooth function to model the whole series the flow becomes the minimal cost \textit{smooth} flow over time. 

\subsection{Further Adaptation for Single-Cell Trajectories}\label{sec:single-cell-data}

Up to this point we have shown how to perform dynamic optimal transport in high dimensions with a regularized CNF. We now introduce priors needed to mimic cellular systems that are characterized by growth/death rather than just transport, endowed with a manifold structure, and knowledge of local velocity arrows. Similar priors may also be applicable to other data types. For example in studying the dynamics of a disease, people may be newly infected or cured, we may have knowledge on acceptable transition states, indicating a density penalty, or visits may be clustered such as in a hospital stay so we may have estimates of near term patient trends, indicating the use of velocity priors. To enforce these priors we add corresponding regularizations listed below:

\begin{enumerate} 

\item A {\em growth rate regularization} that accommodates unbalanced transport, described in Section~\ref{sec:unbalanced}.

\item A {\em density-based penalty} which encourages interpolations that lie on dense regions of the data. Often data lies on a low-dimensional manifold, and it is desirable for paths to follow this manifold at the cost of higher energy (See Section~\ref{sec:density} for details).

\item A {\em velocity regularization} where we enforce local estimates of velocity at measured datapoints to match the first time derivative of cell state change. (See Section~\ref{sec:velocity} for details).
\end{enumerate}

These regularizations are summarized in a single loss function defined as

\begin{equation}\label{eq:tnet_loss}
\begin{split}
    L_T &= \underset{\text{Dynamic OT}}{\underbrace{\overset{\text{Normalizing Flow}}{\overbrace{\sum_{i = 1}^k -\log P_{t_i}(x_{t_i})}} + L_{energy}}} \\
    &+ \underset{\text{Biological priors}}{\underbrace{L_{density} + L_{velocity} + L_{growth}}}
\end{split}
\end{equation}

\subsubsection{Allowing Unbalanced Optimal Transport}\label{sec:unbalanced}  We use a simple and computationally efficient method that adapts discrete static unbalanced optimal transport to our framework in the continuous setting. This is a necessary extension but is by no means a focus of our work. While we could also apply an adversarial framework, we choose to avoid the instabilities of adversarial training and use a simple network trained from the solution to the discrete problem. We train a network $G(x,t) : \mathbb{R}^d \times [0,1] \rightarrow \mathbb{R}^+$, which takes as input a cell state and time pair and produces a growth rate of a cell at that time. This is trained to match the result from discrete optimal transport. For further specification see Appendix~\ref{app:growth-details}. We then fix weights of this network and modify the way we integrate mass over time to
\begin{align}
    \log M_{t_i}(x) 
    &= \log M_{t_{i-1}}(x) 
    - \int_{t_{i-1}}^{t_i} Tr \left ( \frac{\partial f_\theta(x(t),t)}{\partial x(t)} \right ) dt \nonumber \\
    &+ \log G(x_{t_{i-1}},t_{i-1})
     %&+ \sum_{t \in \{t_0, t_1, \ldots, t_{i-1}\}} \log g(x_t,t)
\end{align}
% Where $x_t$ is calculated from $x_{t_i}$ by integration of $f$. 
We note that adding growth rate regularization in this way does not guarantee conservation of mass. We could normalize $M(x)$ to be a probability distribution during training, e.g., as $P(x) = M(x) / \sum_{x \in \mathbb{R}^d} M(x)$. However, this now requires an integration over $\mathbb{R}^d$, which is too computationally costly. Instead, we use the equivalence of the maximum likelihood formulation over a fixed growth function $g$ and normalize it after the network is trained.

\subsubsection{Enforcing Transport on a Manifold}\label{sec:density} Methods that display or perform computations on the cellular manifold often include an implicit or explicit way of normalizing for density and model data geometry. PHATE~\cite{moon_visualizing_2019} uses scatter plots and adaptive kernels to display the geometry. SUGAR~\cite{lindenbaum_geometry_2018} explicitly models the data geometry. We would like to constrain our flows to the manifold geometry but not to its density. We penalize the flow such that it is always close to at least a few measured points across all timepoints.
\begin{align}
    &L_{density}(x, t_d) = \nonumber \\
    &\quad \sum_k \max(0, \text{min-k} \bigl ( \left \{ \left \| x(t_d) - z \right \| : z \in \mathcal{X} \right \} \bigr ) - h)
\end{align}
This can be thought of as a loss that penalizes points until they are within $h$ Euclidean distance of their $k$ nearest neighbors. We use $h=0.1$ and $k=5$ in all of our experiments. We evaluate $L_{density}$ on an interpolated time $t_d \in (t_0, t_k)$ every batch. 

\subsubsection{Conforming to known Velocity}\label{sec:velocity} Often it is the case where it is easy to measure direction of change in a short time horizon, but not have good predictive power at the scale of measured timesteps. In health data, we can often collect data from a few visits over a short time horizon estimating the direction of a single patient in the near future. In single-cell data, RNA-velocity~\cite{la_manno_rna_2018, bergen_generalizing_2019} provides an estimates $\widehat{dx/dt}$ at every measured cell. We use these measurements to regularize the direction of flow at every measured point. Our regularization requires evaluating $f(x,t)$ periodically at every measured cell adding the regularization:
\begin{align}
    L_{velocity}(x, t, \widehat{dx/dt}) 
    &= \text{cosine-similarity}(f(x,t), \widehat{dx/dt}) \nonumber \\
    &= \frac{f(x,t) \cdot \widehat{dx/dt}}{\left \| f(x,t) \right \| \left \| \widehat{dx/dt} \right \|}
\end{align}
This encourages the direction of the flow at a measured point to be similar to the direction of local velocity. This ignores the magnitude of the estimate, and only heeds the direction. While RNA-velocity provides some estimate of relative speed, the vector length is considered not as informative, as it is unclear how to normalize these vectors in a system specific way~\cite{la_manno_rna_2018, bergen_generalizing_2019}. We note that while current estimates of velocity can only estimate direction, this does not preclude future methods that can give accurate magnitude estimates. $L_{velocity}$ can easily be adapted to take magnitudes into account by considering $L_2$ similarity for instance.

\subsection{Training}\label{sec:training}

For simplicity, the neural network architecture of TrajectoryNet consists of three fully connected layers of 64 nodes with leaky ReLU activations. It takes as input a cell state and time and outputs the derivative of state with respect to time at that point. To train a continuous normalizing flow we need access to the density function of the source distribution. Since this is not accessible for an empirical distribution we use an additional Gaussian at $t_0$, defining $P_{t_0}(\cdot) = \mathcal{N}(0,1)$, the standard Gaussian distribution, where $P_t(x)$ is the density function at time~$t$.

For a training step we draw samples $x_{t_i} \sim \mathcal{X}_{t_i} \text{ for } i \in \{1, \ldots, k\}$ and calculate the loss with a single backwards integration of the ODE. In the following sections we will explain how adding the individual penalty terms achieve regularized trajectories. While there are a number of ways to computationally approximate these quantities, we use a parallel method to iteratively calculate the $\log P_{t_i}$ based on $\log P_{t_{i-1}}$. To make a backward pass through all timepoints we start at the final timepoint, integrate the batch to the second to last timepoint, concatenate these points to the samples from the second to last timepoint, and continue till $t_0$, where the density is known for each sample. We note that this can compound the error especially for later timepoints if $k$ is large or if the learned system is stiff, but gives significant speedup during training.

To sample from $P_{t_i}$ we first sample $\hat{x}_{t_0} \sim P_{t_0}$ then use the adjoint method to perform the integration $\hat{x}_{t_i} = \hat{x}_{t_0} + \int_{t_0}^{t_i} f_\theta(x(t), t) dt; \quad x(0) = \hat{x}_{t_0}$.

%\section{Applications to Cell Trajectory Inference in Single Cell Biology}
%    \input{sections/5_applications.tex}
    %\input{figures/Tnet_Circle.tex}
    %\input{figures/Growth_Fig.tex}
\section{Experiments}\label{sec:experiments}
    
All experiments were performed with the TrajectoryNet framework with a network of consisting of three layers with LeakyReLU activations. Optimization was performed on 10,000 iterations of batches of size 1,000 using the dopri5 solver~\cite{dormand_family_1980} with both absolute and relative tolerances set to $1 \times 10^{-5}$ and the ADAM optimizer~\cite{kingma_adam:_2015} with learning rate 0.001, and weight decay $5 \times 10^{-5}$ as in \cite{grathwohl_ffjord:_2019}. We evaluate using three TrajectoryNet models with different regularization terms. The Base model refers to a standard normalizing flow. +E adds $L_{energy}$, +D adds  $L_{density}$,+V adds $L_{velocity}$, and +G adds $L_{growth}$.

\paragraph{Comparison to Existing Methods.}
Since there are no ground truth methods to calculate the trajectory of a single cell we evaluate our model using interpolation of held-out timepoints. We leave out an intermediary timepoint and measure the Kantorovich-distance also known as the earth mover's distance (EMD) between the predicted and held-out distributions. For EMD lower is more accurate. We compare the distribution interpolated by TrajectoryNet with four other distributions. The previous timepoint, the next timepoint, a random timepoint and the McCann interpolant in the discrete OT solution as used in \cite{schiebinger_optimal-transport_2019}.

\begin{figure}[ht]
    \begin{center}
    \includegraphics[width=1 \linewidth]{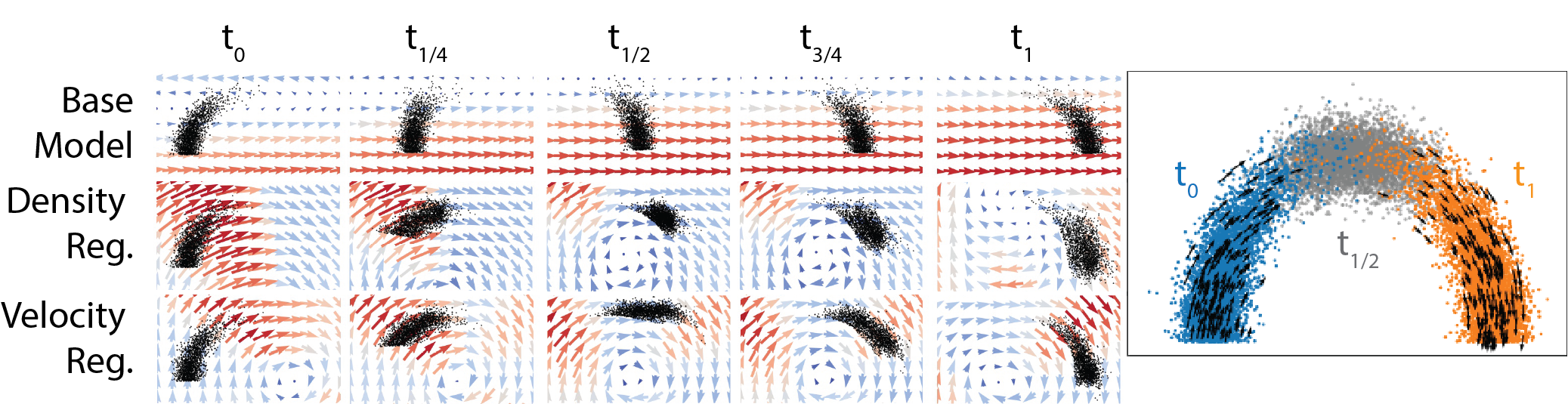}
    \end{center}
    %\vspace{-5mm}
    \caption{Density regularization or velocity regularization can be used to follow a 1D manifold in 2D.}
    \label{fig:tnet-circle}
    %\vspace{-2mm}
\end{figure}
\begin{figure}[ht]
    \begin{center}
    \includegraphics[width=1 \linewidth]{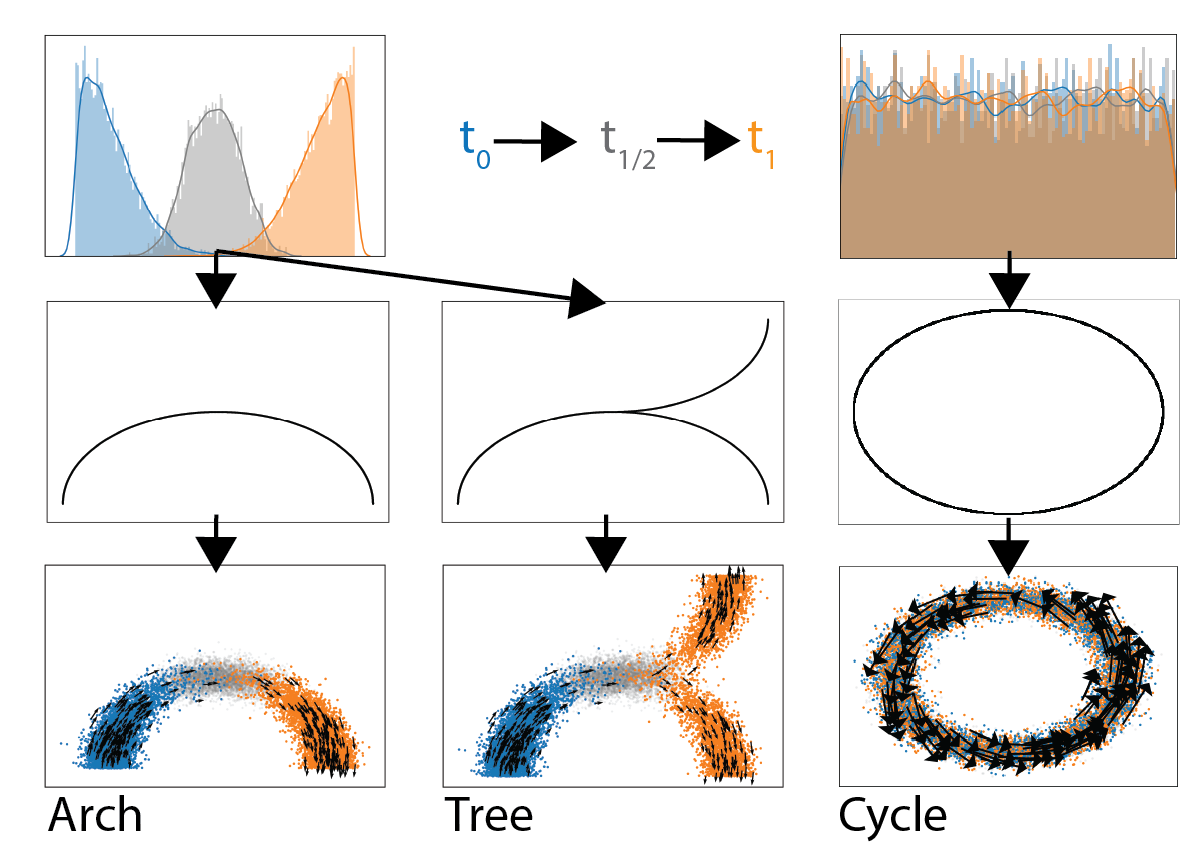}
    \end{center}
    %\vspace{-5mm}
    \caption{A 1D distribution of data over time embedded in two dimensions along a smooth manifold. On a single branch (left), with a tree structure (center), and circle (right).}
    \label{fig:gen-method}
\end{figure}
\begin{table}[hbt]
    \centering
    \scalebox{0.85}{
        \begin{tabular}{lrrrrrr}
\toprule
{} & \multicolumn{3}{l}{EMD} & \multicolumn{3}{l}{MSE} \\
{} &   Arch &  Cycle &   Tree &   Arch &  Cycle &   Tree \\
\midrule
Base         &  0.691 &  0.037 &  0.490 &  0.300 &  0.190 &  0.218 \\
Base + D     &  0.607 &  0.049 &  0.373 &  0.236 &  0.191 &  0.145 \\
Base + V     &  \textbf{0.243} &  0.033 &  \textbf{0.143} &  \textbf{0.107} &  \textbf{0.068} &  \textbf{0.098} \\
Base + D + V &  0.415 &  0.034 &  0.252 &  0.156 &  0.081 &  0.132 \\
\midrule
OT           &  0.644 &  \textbf{0.032} &  0.492 &  0.252 &  0.192 &  0.196 \\
prev         &  1.086 &  0.035 &  1.092 &  0.652 &  0.192 &  0.666 \\
next         &  1.090 &  0.035 &  1.068 &  0.659 &  0.192 &  0.689 \\
rand         &  0.622 &  0.406 &  0.420 &  0.243 &  0.346 &  0.161 \\
\bottomrule
\end{tabular}

    }
    \caption{Shows the Wasserstein distance EMD and MSE for artificial datasets between the left out timepoint and the predicted points for our two generated datasets. Mean over 3 seeds.}
    \label{tab:gen_comparison_table}
    %\vspace{-3mm}
\end{table}
\begin{figure}[ht]
    \begin{center}
    \includegraphics[width=0.95 \linewidth]{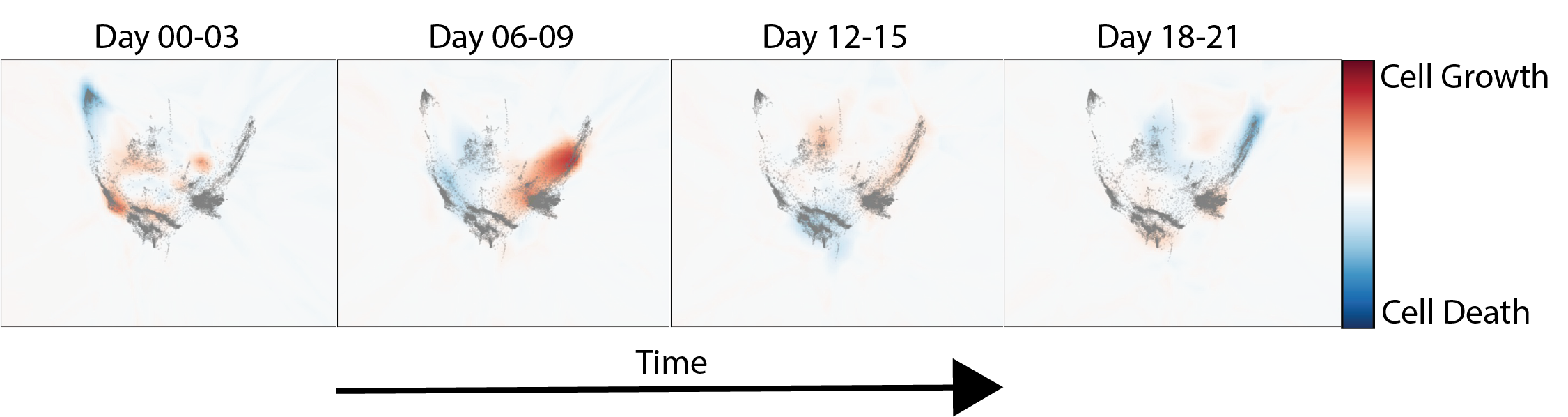}
    \end{center}
    %\vspace{-5mm}
    \caption{Cell growth model learned on Embryoid Body Data~\cite{moon_visualizing_2019}}
    \label{fig:growth-fig}
    \vspace{-5mm}
\end{figure}
\subsection{Artificial Data} 
For artificial data where we have known paths, we can measure the mean squared error (MSE) predicted by the model based on the first timepoint. Here we leave out the middle timepoint $t_{1/2}$ for training then calculate the MSE between the predicted point at time $t_{1/2}$ and the true point at $t_{1/2}$ for 5000 sampled trajectories. This gives a measure of how accurately we can model simple dynamical systems.

We first test TrajectoryNet on two datasets where points lie on a 1D manifold in 2D with Gaussian noise (See Figure~\ref{fig:gen-method}). First two half Gaussians are sampled with means zero and one in one dimension. These progressions are then lifted onto curved manifolds in two dimensions either an arch or a tree mimicking  a differentiating system where we have two sampled timepoints that have some overlap. Table~\ref{tab:gen_comparison_table} shows the Wasserstein distance (EMD) and the mean squared error for different interpolation methods between the interpolated distribution at $t_{1/2}$ and the true interpolated distribution at $t_{1/2}$. Because optimal transport considers the shortest Euclidean distance, the base model and OT methods follow the lowest energy path, which is straight across. With density regularization or velocity regularization TrajectoryNet learns paths that follow the density manifold. Figure~\ref{fig:tnet-circle} and Figure~\ref{fig:tnet-tree} demonstrate how TrajectoryNet with density or velocity regularization learns to follow the manifold.

A third artificial dataset shows the necessity of using velocity estimates for some data. Here we have an unchanging distribution of points distributed uniformly over the unit circle, but are traveling counterclockwise at $\pi / 5$ radians per unit time. This is similar to the cell-cycle process in adult systems. Without velocity estimates it is impossible to pick up this type of dynamical system. This is illustrated by the MSE of the cycle dataset using velocity regularization in Table~\ref{tab:gen_comparison_table}.

\subsection{Single-Cell Data}

We run our model on 5D PCA due to computational constraints, but note that computation time scales roughly linearly with dimension for our test cases (See Appendix~\ref{app:speed}), which is consistent to what was found in \citet{grathwohl_ffjord:_2019}. Since there are no ground truth trajectories in real data, we can only evaluate using distributional distances. We do leave-one-out validation, training the model on all but one of the intermediate timepoints then evaluating the EMD between the validation data and the model's predicted distribution. We evaluate and compare our method on two single-cell RNA sequencing datasets. 

\begin{figure}[ht]
    \begin{center}
    \includegraphics[width=1 \linewidth]{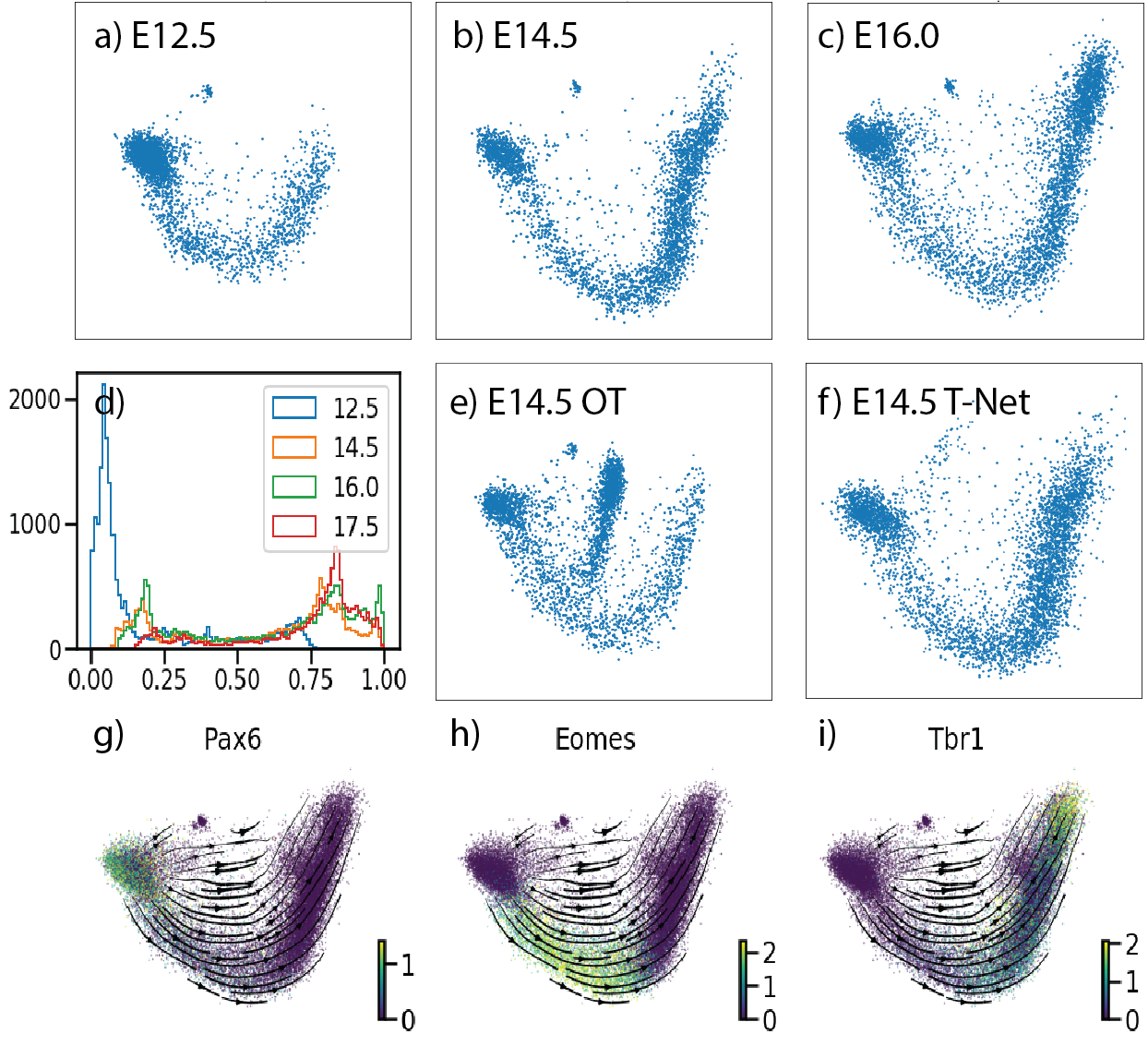}
    \end{center}
    %\vspace{-5mm}
    \caption{Shows the first 2 PCs of the mouse cortex dataset. (a-c) show the distributions for the first three timepoints. (d) shows the distribution of cells over PC1. the interpolated points for E14.5 using (e) static OT, and (f) TrajectoryNet with density regularization. (g-i) shows expression of three markers of early (Pax6) mid (Eomes) and late (Tbr1) stage neurons.}
    \label{fig:mouse-fig}
    \vspace{-5mm}
\end{figure}

\begin{table}[hbt]
\centering
\scalebox{0.89}{
    \begin{tabular}{llll}
\toprule
{} &              rep1 &              rep2 &              mean \\
\midrule
Base         &  0.888 $\pm$ 0.07 &  0.905 $\pm$ 0.06 &  0.897 $\pm$ 0.06 \\
Base + D     &  0.882 $\pm$ 0.03 &  0.895 $\pm$ 0.03 &  0.888 $\pm$ 0.03 \\
Base + V     &  0.900 $\pm$ 0.09 &  0.898 $\pm$ 0.10 &  0.899 $\pm$ 0.10 \\
Base + D + V &  \textbf{0.851} $\pm$ 0.08 &  \textbf{0.866} $\pm$ 0.07 &  \textbf{0.859} $\pm$ 0.07 \\
\midrule
OT           &             1.098 &             1.095 &             1.096 \\
prev         &             1.628 &             1.573 &             1.600 \\
next         &             1.324 &             1.391 &             1.357 \\
rand         &             1.333 &             1.288 &             1.311 \\
\bottomrule
\end{tabular}

}
\caption{Shows the Wasserstein distance between the left out timepoint and the predicted distribution for various methods on a 4 timepoint mouse embryo cortex dataset. Mean and standard deviation over 3 seeds.}
\label{tab:noonan_comparison_table}
%\vspace{-4mm}
\end{table}

\paragraph{Mouse Cortex Data.\footnote{For videos of the dynamics learned by TrajectoryNet see \url{http://github.com/krishnaswamylab/TrajectoryNet}}} The first dataset has structure similar to the Arch toy dataset. It consists of cells collected from mouse embryos at days E12.5, E14.5, E16, and E17.5. In Figure~\ref{fig:mouse-fig}(d)  we can see at this time in development of the mouse cortex the distribution of cells moves from a mostly neural stem cell population at E12.5 to a fairly developed and differentiated neuronal population at E17.5~\cite{cotney_autism-associated_2015, katayama_chd8_2016}. The major axis of variation is neuron development. Over the 4 timepoints we have 2 biological replicates that we can use to evaluate variation between animals. In Table~\ref{tab:noonan_comparison_table}, we can see that TrajectoryNet outperforms baseline models, especially when adding density and velocity information. The curved manifold structure of this data, and gene expression data in general means that methods that interpolate with straight paths cannot fully capture the structure of the data. Since TrajectoryNet models full paths between timepoints, adding density and velocity information can bend the cell paths to follow the manifold utilizing all available data rather than two timepoints as in standard optimal transport.

\begin{figure}[ht]
    \begin{center}
    \includegraphics[width=1 \linewidth]{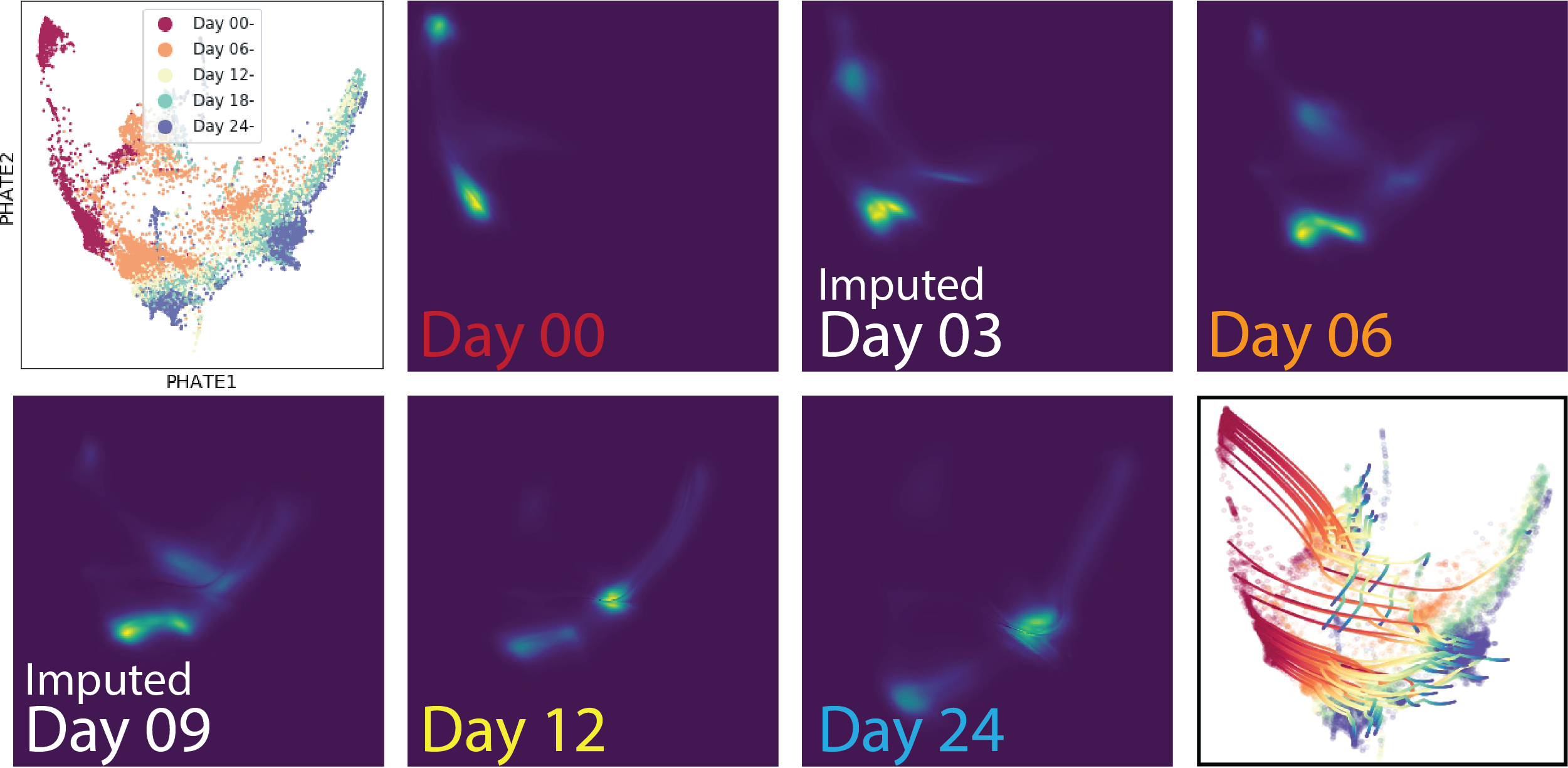}
    \end{center}
    %\vspace{-5mm}
    \caption{Shows the Embryoid body dataset projected into 2D with PHATE~\cite{moon_visualizing_2019} with paths and densities imputed using TrajectoryNet.}
    \label{fig:density-fig}
    %\vspace{-5mm}
\end{figure}

\paragraph{Embryoid body Data.} Next, we evaluate on a differentiating Embryoid body scRNA-seq time course. Figure~\ref{fig:density-fig} shows this data projected into two dimensions using a non-linear dimensionality reduction method called PHATE~\cite{moon_visualizing_2019}. This data consists of 5 timepoints  of single cell data collected in a developing human embryo system (Day 0-Day 24). See Figure~\ref{fig:growth-fig} for a depiction of the growth rate. Initially, cells start as a single stem cell population, but differentiate into roughly 4 cell precursor types. This gives a branching structure similar to our artificial tree dataset. In Table~\ref{tab:eb_table} we show results when each of the three intermediate timepoints are left out. In this case velocity regularization does not seem to help, we hypothesis this has to do with the low unspliced RNA counts present in the data (See Figure~\ref{fig:eb-unspliced-counts}). We find that energy regularization and growth rate regularization help only on the first timepoint, and that density regularization helps the most overall.

\begin{figure}[ht]
    \begin{center}
    \includegraphics[width=1 \linewidth]{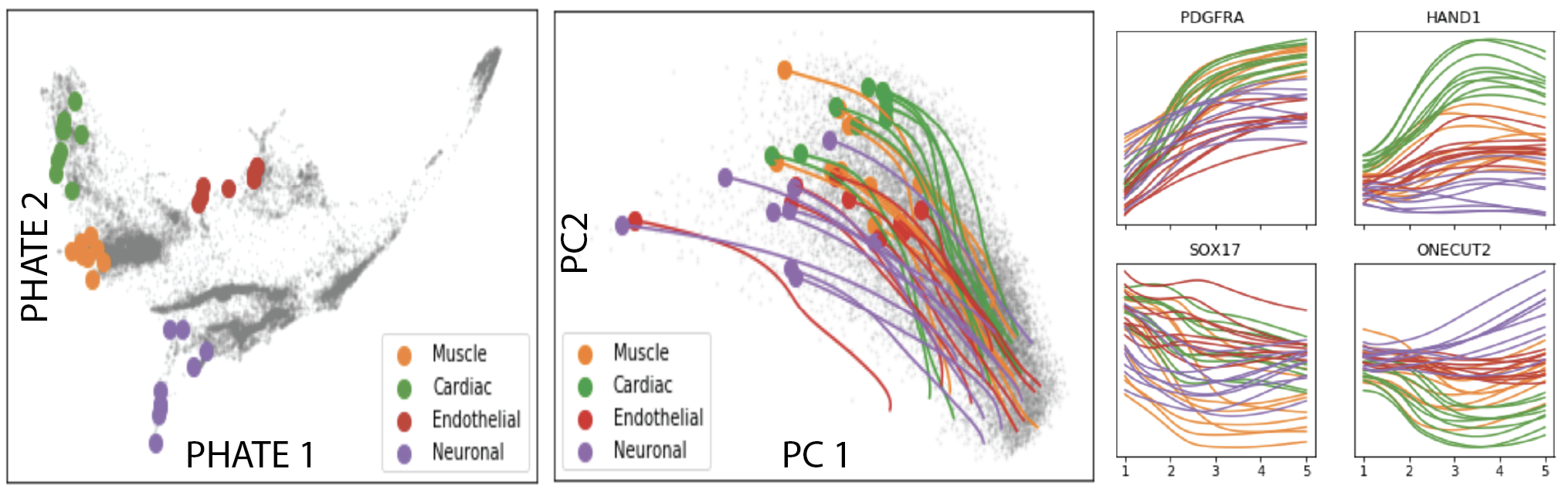}
    \end{center}
    %\vspace{-5mm}
    \caption{For curated endpoints, shows location on PHATE dimensions, TrajectoryNet paths projected into PCA space, and trajectories for 4 genes.}
    \label{fig:mouse-genes}
    %\vspace{-5mm}
\end{figure}

We can also project trajectories back to gene space. This gives insights into when populations might be distinguishable. In Figure~\ref{fig:mouse-genes}, we demonstrate how TrajectoryNet can be projected back to the gene space. We sample cells from the end of the four main branches, then integrate TrajectoryNet backwards to get their paths through gene space. This recapitulates known biology in \citet{moon_visualizing_2019}. See appendix~\ref{app:genes} for a more in-depth treatment.

\begin{table}[hbt]
    \centering
    \scalebox{1}{
\begin{tabular}{lrrrr}
\toprule
{} &    t=1 &    t=2 &    t=3 &   mean \\
\midrule
Base         &  0.764 &  0.811 &  0.863 &  0.813 \\
Base + D     &  0.759 &  \textbf{0.783} &  0.811 &  \textbf{0.784} \\
Base + V     &  0.816 &  0.839 &  0.865 &  0.840 \\
Base + D + V &  0.930 &  0.806 &  \textbf{0.810} &  0.848 \\
Base + E     &. 0.737 &  0.896 &  0.842 &  0.825 \\
Base + G     &  \textbf{0.700} &  0.913 &  0.829 &  0.814 \\
\midrule
OT           &  0.791 &  0.831 &  0.841 &  0.821 \\
prev         &  1.715 &  1.400 &  0.814 &  1.309 \\
next         &  1.400 &  0.814 &  1.694 &  1.302 \\
rand         &  0.872 &  1.036 &  0.998 &  0.969 \\
\bottomrule
\end{tabular}

}
 \caption{Shows the Wasserstein distance (EMD) between the left out timepoint and the predicted distribution for various methods on the 5 timepoint Embryoid body dataset.}
    \label{tab:eb_table}
    
    %\vspace{-3mm}
\end{table}
 
%\paragraph{Improvements over the basic model}
%A more discriminative evaluation uses negative log likelihoods (NLL) to avoid convergence problems of the Wasserstein distance. Here we compare our base model with the addition of variable growth rates, velocity, and density regularization. We find that adding these factors leads to a more accurate model of cell distribution interpolation in some instances. Adding the growth model leads to worse fit in the first few timepoints, but a better fit in later timepoints. Overall, we find that the additional regularizations do not change the ability of the model to fit the data, but only the paths taken between distributions. Figure~\ref{fig:training} shows the training negative log likelihood for various additions to the base model. The regularizations do not change the overall training fit very much, but do change the transport paths (see Figure~\ref{fig:density-fig}). 

\section{Conclusion}
    %TrajectoryNet demonstrates how neural networks can be used as a probabilistic model for continuous trajectory inference. 

TrajectoryNet computes dynamic optimal transport between distributions of samples at discrete times to model realistic paths of samples continuously in time. In the single-cell case, TrajectoryNet "reanimates," cells which are destroyed by measurement to recreate a continuous-time trajectory. This is also relevant when modeling any underlying system that is high-dimensional, dynamic, and non-linear. In this case, existing static OT methods are under-powered and do not interpolate well to intermediate timepoints between measured ones. Existing dynamic OT methods (non-neural network based) are computationally infeasible for this task. 

In this work we integrate multiple priors and assumptions into one model to bias TrajectoryNet towards more realistic dynamic optimal transport solutions. We demonstrated how this gives more power to discover hidden and time specific relationships between features. In future work, we would like to consider stochastic dynamics~\cite{li_scalable_2020} and learning the growth term together with the dynamics.

%Because of computational constraints our model is deterministic, given a pair $(x,t)$ the trajectory through time is deterministic. While there is recent work in extending the neural ODE framework to the stochastic regime~\cite{liu_neural_2019} these works rely on stochastic differential equation solvers, which converge too slowly for our applications. Furthermore, since we use an ODE solver, we cannot easily handle the unbalanced optimal transport problem~\cite{yang_scalable_2019} \todo{More citations here chizat? Computation OT book} as defined for the dynamic optimal transport problem in chizat et al. \todo{cite} as their formulation requires solving a PDE. 

%While limiting our solver to ODEs has some limitations it is computationally efficient and able to scale to large and practical problems such as those in the cell lineage tracking problem.

\section*{Acknowledgements}
This research was partially funded by IVADO (l'institut de valorisation des donn\'{e}es) [\emph{G.W.}]; Chan-Zuckerberg Initiative grants 182702 \& CZF2019-002440 [\emph{S.K.}]; and NIH grants R01GM135929 \& R01GM130847 [\emph{G.W., S.K.}]. The content provided here is solely the responsibility of the authors and does not necessarily represent the official views of the funding agencies.

\bibliography{main,auto_DO_NOT_MODIFY}
\bibliographystyle{icml2020}

\clearpage
\appendix
\beginsupplement
\section*{Supplement}
\section{Technical details}

\subsection{Proof of Theorem~\ref{thm:approach}}\label{app:proof}
% We now show that the solution of \ref{eq:wasserstein} for sufficiently large $\lambda$ converges to the optimal solution of equation~\ref{eq:benamou} when subject to the conditions ~\ref{eq:benamou:c2}. 

First, we apply the Lagrange multiplier method by introducing the variable $\lambda$ to the minimization problem of~\eqref{eq:benamou} subject to constraints~\eqref{eq:benamou:c2}. As we always begin with the base distribution and at any time $t$, x is defined by $f(x,t)$ and the initial value $x(t_0)=x_0$, which has $\text{KL}(P(t_0,\cdot)\parallel\mu) = 0$. 

\begin{equation}
\begin{split}
     \inf_{(P, f)} \sup_{\lambda} \left(t_1-t_0\right)\mathop{\mathbb{E}}_{x_0 \sim \mu} \int_{t_0}^{t_1} & P(x,t) \left|f(x,t)\right|^2 dt\\
    &+ \lambda \text{KL}(P(t_1, \cdot) \parallel \nu)
    % \inf_{(P, f)} \sup_{\lambda} \left(t_1-t_0\right)\int_{\mathbb{R}^d} \int_{t_0}^{t_1} & P(x,t) \left|f(x,t)\right|^2 dt dx \\
    % &+ \lambda \text{KL}(P(t_1, \cdot) \parallel \nu)
\end{split}
\end{equation}

The expectation part is equivalent to  $\int_{\mathbb{R}^d} \int_{t_0}^{t_1} P(x,t) \left|f(x,t)\right|^2 dt dx $, and we use interchangeably in the the remaining of the proof. Since the KL divergences are non-negative, $\lambda \geq 0$. The optimal solution of the min-max problem is the optimal solution to the original problem. Consider the true minimal loss given by the optimal solution to be $c$, we know that $L(\lambda)\leq c$, where

\begin{equation}\label{eq:max-min}
\begin{split}
    L(\lambda) = \sup_{\lambda \geq 0} \inf_{(P, f)} & \left(t_1-t_0\right)\mathop{\mathbb{E}}_{x_0 \sim \mu} \int_{t_0}^{t_1} P(x,t) \left|f(x,t)\right|^2 dt\\
     &  + \lambda \text{KL}(P(t_1, \cdot) \parallel \nu)
    %  L(\lambda) = \sup_{\lambda \geq 0} \inf_{(P, f)} & \left(t_1-t_0\right)\int_{\mathbb{R}^d} \int_{t_0}^{t_1} P(x,t) \left|f(x,t)\right|^2 dt dx \\
    %  &  + \lambda \text{KL}(P(t_1, \cdot) \parallel \nu)
\end{split}
\end{equation}

In order to show that the solution of the max-min problem converges to $c$, we first show that it is monotonic in $\lambda$. For easier reading, set $E = \int_{\mathbb{R}^d} \int_{t_0}^{t_1} P(x,t) \left|f(x,t)\right|^2 dt dx $, $M = \text{KL}(P(t_1, \cdot) \parallel \nu) $. Both $E$ and $M$ are functions of $f$. For any pair of $E, M$ values, if $\lambda_1 > \lambda_2$, $E+\lambda_1 M > E+ \lambda_2 M$. Thus the maximum and minimum of the function $L = E + \lambda M$ is also monotonic in $\lambda$, and it will converge to the supremum. 

Next, we show that the divergence term $M(f)$ decreases monotonically as $\lambda$ increases, and it converges to $0$ as $\lambda$ goes to infinity. For a given $\lambda$, let $f_\lambda ^*  = \text{arg}\inf E(f) + \lambda M(f)$.  By definition, $E(f^*_{\lambda_1}) + \lambda_1 M(f^*_{\lambda_1}) \leq E(f^*_{\lambda_2}) + \lambda_1 M(f^*_{\lambda_2}) $, and $E(f^*_{\lambda_1}) + \lambda_2 M(f^*_{\lambda_1}) \geq E(f^*_{\lambda_2}) + \lambda_2 M(f^*_{\lambda_2}) $. Thus $(\lambda_1 - \lambda_2)(M(f^*_{\lambda_1}) -M(f^*_{\lambda_2})) \leq 0$. If $\lambda_1 > \lambda_2$,then $M(f^*_{\lambda_1}) -M(f^*_{\lambda_2}) \leq 0$. The sequence $M(f)$ decreases monotonically as $\lambda$ increases. Because $L(\lambda)$ is upper bounded by $c$ and $M(f)\geq 0$, $M(f)$ converges to zero as $\lambda$ goes to infinity.

Now we have shown that $L(\lambda)$ is a monotone sequence and is upper bounded by $c$, and that the divergence term $M$ converges to zero, we next show that $L$ converges to $c$, and that the optimal solution of the max-min problem in~\eqref{eq:max-min} is the optimal solution of the original problem. Since the divergence term is non-negative, we have a lower bound for $L(\lambda)$ as 

\begin{equation}
    \begin{split}
        \forall \lambda, \quad  L(\lambda) & \geq \inf E(f) \\
        & \quad  \text{s.t.} \quad M(f) \leq M(f^*_\lambda)
    \end{split}
\end{equation}

Because $L(\lambda)$ is monotonically increasing, and $M(f)$ is monotonically decreasing, $H(\lambda) =\inf E(f)$ increases as $\lambda$ increases. 

\begin{lemma}\label{lemma:E-diff}
    
        \begin{align*}
            \forall \epsilon > 0, \quad \forall f, \quad s.t. & \quad M(f) \leq \epsilon,  \\
            \exists \hat{f}, \quad s.t. & \quad M(\hat{f}) = 0, \quad {\text and} \\
             E (\hat{f}) - E(f)  \leq  & \frac{D^2}{\sqrt{2}T}\sqrt[\leftroot{-2}\uproot{2}4]{\epsilon}(1+\sqrt[\leftroot{-2}\uproot{2}4]{\epsilon})
        \end{align*}
\end{lemma}

where $D$ is the diameter of the probability space and $T$ is the transformation completion time.

\begin{proof}
For a certain $\lambda$, starting from the base distribution $\mu$, at time $T$ the distribution is transformed, by following $f$, to $\nu'$, and KL$(\nu' \parallel \nu) = \epsilon$. Now consider a different transformation $\hat{f}$, which is composed of two part:  the first part is an accelerated $f$, so that $\nu'$ is achieved by time $T/(1+\xi)$, and the second part is transforming $\nu'$ to $\nu$ in the remaining time of $\frac{\xi T}{(1+\xi)}$. Thus at time $T$, by following $\hat{f}$, we achieve zero divergence, $M(\hat{f})=0$. The new transformation $\hat{f}$ has an increased $E$, from the acceleration and the additional transformation. 

\begin{equation}\label{eq:E-fhat}
    \begin{split}
        E (\hat{f})  = & \frac{T}{1+\xi}\int_{\mathbb{R}^d} \int_{0}^{\frac{T}{1+\xi} } P\left(x,\left(1+\xi\right)t\right)\\
        & \quad \quad \quad \quad \quad |(1+\xi)f(x,(1+\xi)t)|^2 dt dx \\
        & + \int_{Z} \int_{Y} \int_{\frac{T}{1+\xi}}^{T}  \mathcal{M}(y,z) \frac{|z-y|^2}{(\frac{\xi}{1+\xi}T)^2} dt dy dz 
    \end{split}
\end{equation}

where $Z$ has distribution $\nu$, $Y$ has distribution $\nu'$, and $\mathcal{M}(y,z)$ is a mapping from $Y$ to $Z$. The first part of $E(\hat{f})$ is just $E(f)$. The second part is upper bounded by $\frac{TV(Y,Z)D^2}{\frac{\xi}{1+\xi}T}$, where $TV(Y,Z)$ is the total variation between $Y$ and $Z$, which is in turn upper bounded by $\sqrt{\epsilon/2}$. We choose $\xi =\sqrt[\leftroot{-2}\uproot{2}4]{\epsilon} $. 
\end{proof}

By definition, $E(\hat{f}) \geq c$, as $c$ is the infimum at zero divergence. Assuming $L(\lambda)$ converges to $c-\alpha$, and $\alpha >0$, $c-\alpha \geq E(f)$.  Then by Lemma~\ref{lemma:E-diff}, $\forall \epsilon, \quad \alpha <  \frac{D^2}{\sqrt{2}T}\sqrt[\leftroot{-2}\uproot{2}4]{\epsilon}(1+\sqrt[\leftroot{-2}\uproot{2}4]{\epsilon})$, and we have a contradiction. Now we complete the proof that $L(\lambda)$ converges to $c$ and that for a large enough $\lambda$, the solution of the max-min problem~\eqref{eq:max-min} is the solution of~\eqref{eq:benamou} when subject to conditions~\eqref{eq:benamou:c2}.

% We now show how relaxing the constrain in equation~\ref{eq:benamou:c2} is valid. For a full proof see~\cite{liero_optimal_2018}. 

% First we revisit equation~\ref{eq:benamou} when subject to the conditions ~\ref{eq:benamou:c2}, we first relax the constraints by introducing divergence as following:

% \begin{align}
% \inf_{(\rho, v)} (t_1 - t_0) \int_{\mathbb{R}^d} \int_{t_0}^{t_1} \rho(x,t) |v(x,t)|^2 dx dt\\
% \text{s.t.} \quad  D_\psi(\rho(t_1, \cdot) \mid \rho_1) +  D_\psi(\rho(t_0, \cdot) \mid \rho_0) \leq c,
% \end{align}

% where $c$ is a parameter of choice. Thus we can have the primal as: 
% \begin{equation}
%     \inf_{(\rho, v)} \sup_{\lambda \geq 0} \int_{\mathbb{R}^d} \int_{t_0}^{t_1} \rho(x,t) |v(x,t)|^2 dx dt + \lambda (D_\psi(\rho(t_1, \cdot) \mid \rho_1) +  D_\psi(\rho(t_0, \cdot) \mid \rho_0) - c)
% \end{equation}

% Assuming convexity, we have the dual as
% \begin{equation}
%     \sup_{\lambda \geq 0} \inf_{(\rho, v)} \int_{\mathbb{R}^d} \int_{t_0}^{t_1} \rho(x,t) |v(x,t)|^2 dx dt + \lambda (D_\psi(\rho(t_1, \cdot) \mid \rho_1) +  D_\psi(\rho(t_0, \cdot) \mid \rho_0) - c).
% \end{equation}

% Since we start with samples from the base distribution $\rho_0$, the corresponding divergence is zero, set $c=0$, and our target becomes:
% \begin{equation}
%     \sup_{\lambda \geq 0} \inf_{(\rho, v)} \int_{\mathbb{R}^d} \int_{t_0}^{t_1} \rho(x,t) |v(x,t)|^2 dx dt + \lambda D_\psi(\rho(t_1, \cdot) \mid \rho_1).
% \end{equation}

%\section{Single Cell Details}

\section{Growth Rate Model Training}\label{app:growth-details}
Our growth network $G(x,t)$ is trained to match the discrete unbalanced optimal transport problem with entropic regularization:
\begin{align}
    \gamma 
    &= \text{argmin}_\gamma < \gamma, M >_F 
    + \lambda \bigl ( \sum_{i,j} \gamma_{i,j} \log (\gamma_{i,j}) \bigr ) \nonumber \\
    &+ \alpha KL(\gamma 1, \mu) + \beta KL(\gamma^T 1, \nu) \\
    &\text{s.t. } \gamma \ge 0 \nonumber
\end{align}
Where $< \cdot, \cdot >_F$ is the Frobenius norm of elementwise matrix multiplication of the transportation matrix $\gamma$ and the cost matrix $M$, and where $\lambda, \alpha, \beta$ are regularization constants $>0$ on the source and target unbalanced distributions $\mu, \nu$. Then the growth rate of each cell $i$ in $\mu$ to $\nu$ is then

\begin{equation}
    g_i = \gamma_{(i, \cdot)} 1
\end{equation}

In our experiments we set $\lambda = 0.1, \beta = 10000$ and tune $\alpha$ for reasonable growth rates. This gives a growth rate at every observed cell; however, our model needs a growth rate defined continuously at every measured timepoint. For this, we learn a neural network that is trained to match the growth rate at measured cells and equal to one at negative sampled points. We use a simple form of negative sampling, for each batch of real points we sample an equal sized batch of points from a uniform distribution over the [-1,1] hypercube, where these negative points are given a growth rate value of 1. The network is trained with mean squared error loss to match these growth rates at all measured times.

\section{Scaling with Dimension}\label{app:speed}

\begin{figure}[ht]
    \begin{center}
    \includegraphics[width=1 \linewidth]{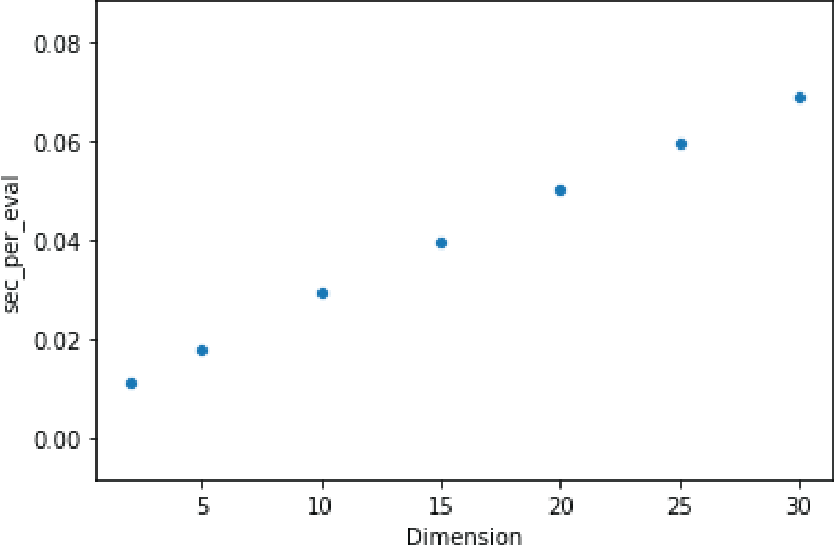}
    \end{center}
    \caption{The computation per evaluation is roughly linear in terms of dimension.}
    \label{fig:time_v_dim}
\end{figure}

\paragraph{Runtime Considerations.} Existing numerical methods for solving dynamic OT rely on proximal splitting methods over a discretized staggered grid~\cite{benamou_computational_2000, papadakis_optimal_2014, peyre_computational_2019}. This results in a non-smooth but convex optimization problem over these grid points. However, the number of grid points scales exponentially with the dimension, so these methods are only applicable in low dimensions. TrajectoryNet scales polynomially with the dimension. See Figure~\ref{fig:time_v_dim} for empirical measurements.

To test the computation time with dimension we run TrajectoryNet for 100 batches of 1000 points on the mouse cortex dataset over different dimensionalities. For hardware we use a single machine with An AMD Ryzen Threadripper 2990WX 32-core Processor, 128GB of memory, and three Nvidia TITAN RTX GPUs. Our model is coded in the Pytorch framework~\cite{paszke_pytorch_2019}.
We count the total number of function evaluations (both forward and backward) divide the total time by this. In Figure~\ref{fig:time_v_dim}, you can see the seconds per evaluation is roughly linear with the dimensionality of the data. This does not imply convergence of the model is linear in dimension, only that computation per iteration is linear. As suggested in \citet{grathwohl_ffjord:_2019}, number of iterations until convergence is a function of how complicated the distributions are, and less dependent on the ambient dimension itself. By learning flows along a manifold with $L_{density}$, our method may scale closer to the intrinsic dimensionality of the data rather than the ambient.

\begin{figure*}[ht]
    \begin{center}
    \includegraphics[width=1 \linewidth]{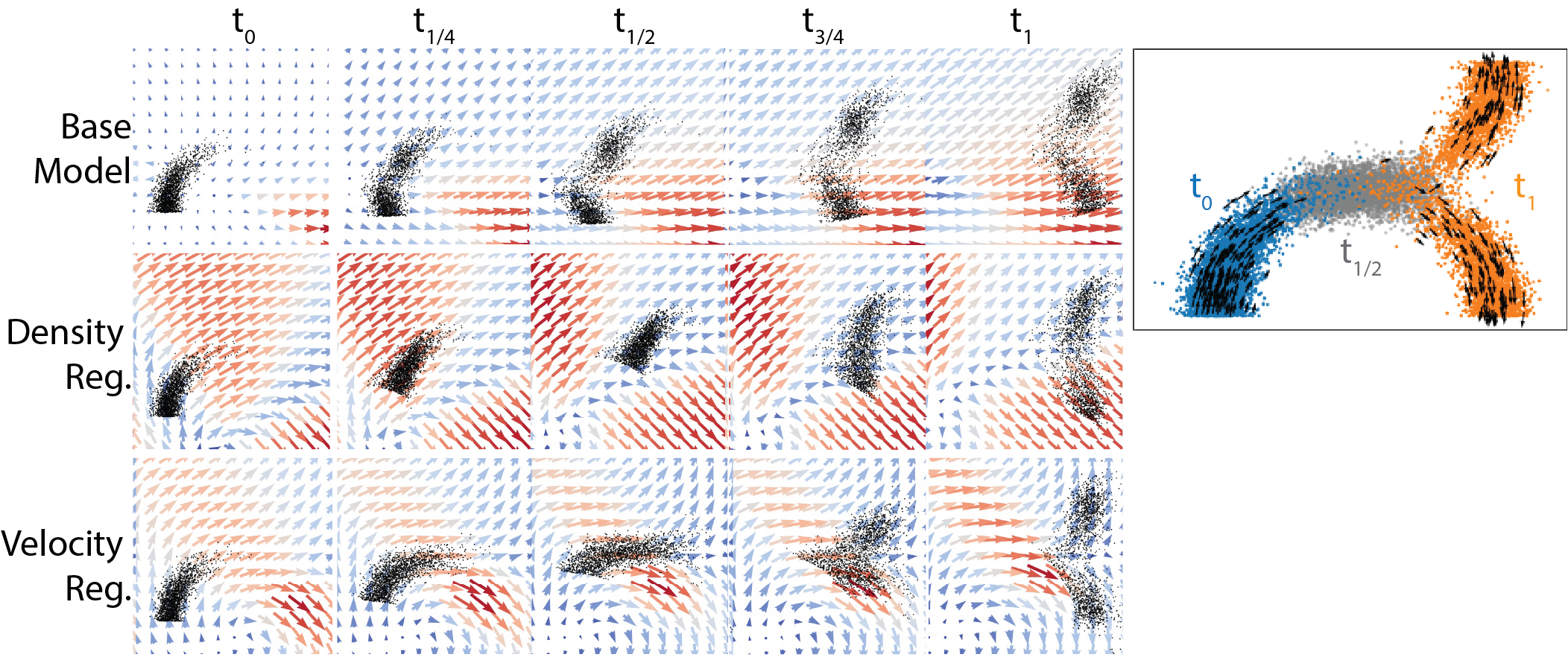}
    \end{center}
    \caption{Density regularization or velocity regularization can be used to follow a 1D manifold in 2D.}
    \label{fig:tnet-tree}
\end{figure*}

\section{Biological Considerations}\label{app:genes}
\begin{figure}[ht]
    \begin{center}
    \includegraphics[width=0.5 \linewidth]{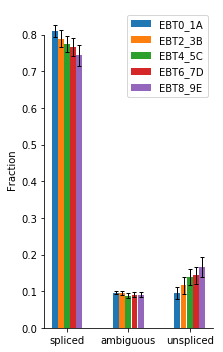}
    \end{center}
    \caption{Shows the ratio of spliced, ambiguous, and unspliced RNA counts over the 5 timepoints in the Embryoid body dataset. Mean unspliced here is around $10\%$-$20\%$ of total counts, in other systems this is near $30\%$~\cite{la_manno_rna_2018}.}
    \label{fig:eb-unspliced-counts}
\end{figure}

Quality control and normalization is important when estimating RNA-velocity from existing single cell measurements. We suspect that the RNA-velocity measurements from the Embryoid body data may be suspect given the low number of unspliced RNA counts present. In Figure~\ref{fig:eb-unspliced-counts} we can see that each timepoint consists of around $10\%$-$20\%$ of unspliced RNA. This is relatively low relative to numbers in other recent works~\cite{la_manno_rna_2018, bergen_generalizing_2019, kanton_organoid_2019}. Low unspliced RNA counts leads to more noise in the estimates of RNA velocity and lower quality. 

In Figure~\ref{fig:mouse-genes} we showed how TrajectoryNet can be projected back to the gene space. These projections can be used to infer the differences much earlier in time than they can be identified in the gene space. Here we have four populations that are easily identified by marker genes or clustering at the final timepoint. Since all four populations emerge from a single relatively uniform stem cell population, the question becomes how early can we identify the features of progenitor cells, the cells leading to these differentiated populations. Since TrajectoryNet models cells as probabilities continuously over time, we can find the path for each differentiated cell in earlier timepoints. This allows inferences such as the fact that HAND1, a gene that is generally high in cardiac cells, is high at earlier timepoints, and may even start to distinguish the population as early as day 6. A gene like ONECUT2 is only starts to distinguish neuronal populations at later timepoints. For further information on this particular system see \cite{moon_visualizing_2019} Figure 6. 

\section{Reproducibility}\label{app:reproducibility}

To foster reproducibility, we provide as many details as possible on the experiments in the main paper. Code is available at \url{http://github.com/krishnaswamylab/TrajectoryNet}.
\subsection{2D Examples}
In Figure~\ref{fig:scurve} we transport a Gaussian to an s-curve. The Gaussian consists of 10000 points sampled from a standard normal distribution. The s-curve is generated using the sklearn function \texttt{sklearn.datasets.make\_s\_curve} with noise of 0.05, and 10000 samples. We then take the first and third dimension, and multiply by 1.5 for the proper scaling. To generate the OT subplot we used the Mccann interpolant from 200 points sampled from the Gaussian. To generate panel (d), we used the procedure detailed in the beginning of Section~\ref{sec:experiments} to train TrajectoryNet, then sampled 200 points from a Gaussian distribution and used the adjoint with these points as the initial state at time $t_0$ to generate points at time $t_1$. For panel (e) we added an energy regularization with $\lambda_e=0.1$ and $\lambda_j=1$. These were found by experimentation, although parameters in the range of $\lambda_e=[0.01, 1]$ and $\lambda_j=[0.1,1]$ were largely visually similar.

To generate the arch and tree datasets we started with two half Gaussians $\mathcal{N}(\cdot, \frac{1}{2 \pi})$ at mean zero and one (as pictured in Figure~\ref{fig:gen-method}) with 5000 points each, then found the Mccann interpolant at $t_{1/2}$ as the test distribution. We then lift these into 2d by embedding on the half circle of radius 1 and adding noise $\mathcal{N}(0,0.1)$ to the radius. To generate velocity, we add a velocity tangent to the circle for each point. For the tree dataset we additionally flip (randomly) half of the points with $x>1$ over the line $y=1$. 

For the Cycle dataset, we start with 5000 uniformly sampled points around the circle, with radius as $\mathcal{N}(1,0.1)$ We then add an arrow tangent to the circle with magnitude $\pi/5$, Thus in one time unit the points should move $1/10$ of the way around the circle. 

\subsection{Single Cell Datasets}
Both single cell datasets were sequenced using 10X sequencing. The Embryoid body data can be found here\footnote{\url{https://doi.org/10.17632/v6n743h5ng.1}} and consists of roughly 30,000 cells unfiltered, and 16,000 cells after filtering. The mouse cortex dataset is not currently publicly available, but consists of roughly 20,000 cells after filtering. For both datasets no batch correction was used. Raw sequences were processed with CellRanger. We then used velocyto~\cite{la_manno_rna_2018} to produce the unspliced and spliced count matrices. We then used the default parameters in  ScVelo~\cite{bergen_generalizing_2019} to generate velocity arrows on a PCA embedding. These include count normalization across rows, selection of the 3000 most variable genes, filtering of low quality genes, and smoothing counts between cells. 

For parameters we did a grid search over $\lambda_{density} \in \{0, 0.1, 0.01\}$, $\lambda_{velocity} \in \{0, 0.001, 0.0001\}$. For Base+E we did a search of $\lambda_{energy} \in \{1.0, 0.1, 0.01\}$, A more extensive search could lead to better results. We intended to show how these regularizations can be used and demonstrate the viability of this approach rather than fully explore parameter space.

\subsection{Software Versioning}
The following software versions were used.
\texttt{scvelo==0.1.24, torch==1.3.1, torchdiffeq==0.0.1, velocyto==0.17.17, scprep==1.0.3, scipy==1.4.1, scikit-learn==0.22, scanpy==1.4.5
}

\end{document}